\documentclass[letterpaper, 10 pt, conference]{ieeeconf}  

\IEEEoverridecommandlockouts                              

\usepackage{amsmath}
\usepackage{amssymb}
\usepackage{dsfont}
\usepackage{float}
\usepackage{algpseudocode}
\usepackage{multirow,booktabs,color,soul}
\newif\ifsubmissionmode
\submissionmodetrue

\ifsubmissionmode
\usepackage[draft,bookmarks=false]{hyperref}
\else
\usepackage[colorlinks=true,linkcolor=blue,citecolor=blue,urlcolor=blue]{hyperref}
\makeatletter
\def\@citex[#1]#2{%
  \leavevmode
  \let\@citea\@empty
  \@cite{%
    \@for\@citeb:=#2\do{%
      \@citea\def\@citea{], [}%
      \edef\@citeb{\expandafter\@firstofone\@citeb\@empty}%
      \if@filesw\immediate\write\@auxout{\string\citation{\@citeb}}\fi
      \@ifundefined{b@\@citeb}{%
        \mbox{\reset@font\bfseries?}%
        \G@refundefinedtrue
        \@latex@warning{Citation `\@citeb' on page \thepage\space undefined}%
      }{%
        \edef\@cite@num{\csname b@\@citeb\endcsname}%
        \hyper@linkstart{cite}{Item.\@cite@num}\@cite@num\hyper@linkend
      }%
    }%
  }{#1}%
}
\makeatother
\fi
\usepackage{graphicx}
\usepackage{cleveref}
\Crefname{figure}{Fig.}{Figs.}
\crefname{figure}{fig.}{figs.}
\usepackage{booktabs}
\usepackage{subfigure}
\usepackage{multirow}
\usepackage[absolute]{textpos}
\textblockorigin{1in}{1in}
\usepackage{siunitx}
\AtBeginDocument{%
  \abovedisplayskip=6pt plus 2pt minus 2pt
  \belowdisplayskip=6pt plus 2pt minus 2pt
  \abovedisplayshortskip=3pt plus 1pt
  \belowdisplayshortskip=3pt plus 1pt
}
\let\OLDthebibliography\thebibliography
\renewcommand\thebibliography[1]{%
  \OLDthebibliography{#1}%
  \setlength{\parskip}{0pt}%
  \setlength{\itemsep}{0pt plus 0.3ex}%
}

\usepackage{bm}
\usepackage{colortbl}
\definecolor{mygray}{gray}{.9}

\title{\LARGE \bf
Disturbance-Aware Flight for Aerial Robots in Narrow Space
}

\author{
	Lei Qiang, Tianyu He, Chenyang Sun, Xurui Liu, Miao Wang, Xiaobin Zhou\textsuperscript{\dag}
	\thanks{
	This work was supported in part by the National Natural Science Foundation of China 
	under Grant 62303412, in part by the Natural Science Foundation of Huzhou City, 
	Zhejiang Province, under Grant 2023YZ01, and in part by the State Key Laboratory 
	of Advanced Rail Autonomous Operation (Project No. RAO2026K09), Beijing Jiaotong University.
	(Corresponding author: Xiaobin Zhou)
	}
	\thanks{
	The authors are with the School of Robotics and Automation, Nanjing University,
	Suzhou 215163, China.
	(Emails: qianglei1204@gmail.com, tonyhe\_2015321@outlook.com, 502025820007@smail.nju.edu.cn, xrl250444@gmail.com, wm\_ubuntu@nju.edu.cn, xiaobin\_nju@nju.edu.cn)
	}
}

\begin{document}

\maketitle
\thispagestyle{empty}
\pagestyle{empty}

\begin{abstract}
	Autonomous flight of aerial robots in narrow space remains challenging due to strong aerodynamic disturbances and limited flying space. Existing approaches mainly address aerodynamic disturbances at the control level, while motion planning typically relies on geometric constraints and fixed speed limits, leading to conservative or unsafe behaviors in confined environments. This paper presents a disturbance-aware planning and control framework (DAPCF) that integrates online disturbance estimation into the planning-control loop for quadrotor flight in narrow space. First, the dual-loop observers estimate 6-degree-of-freedom disturbance forces and torques in real time based on odometry and motor speed measurements. Then, a disturbance risk function is introduced that adaptively modulates the reference speed of the planner based on disturbance estimation, reducing velocity when disturbances exceed a threshold and restoring it under low-disturbance conditions. Finally, a motor-dynamics-based nonlinear model predictive controller (MDNMPC) with disturbance compensation is designed to ensure robust trajectory tracking under perturbed conditions. Experiments demonstrate that a quadrotor with a diagonal length of 0.39~m can traverse straight, sloped, and curved tunnels as narrow as 0.6~m, outperforming human pilots in both success rate and flight efficiency.
\end{abstract}

\section{INTRODUCTION}

Multirotor aerial robots have shown great potential in applications such as search and rescue, infrastructure inspection, transportation, and surveillance due to their agility and ability to access hazardous or confined environments~\cite{sun2025agile, loianno2016estimation, zhou2026trirotor}. In many real-world scenarios, aerial robots are required to operate inside narrow space such as pipelines, tunnels, underground corridors, and damaged buildings. However, reliable autonomous flight in such confined environments remains highly challenging. The extremely limited free space significantly reduces collision margins~\cite{oleynikova2017safe}, while aerodynamic proximity effects caused by nearby walls, grounds, and ceilings introduce complex disturbances that degrade flight stability and control performance.

\begin{figure}
	\centering
	\includegraphics[width=0.990\linewidth]{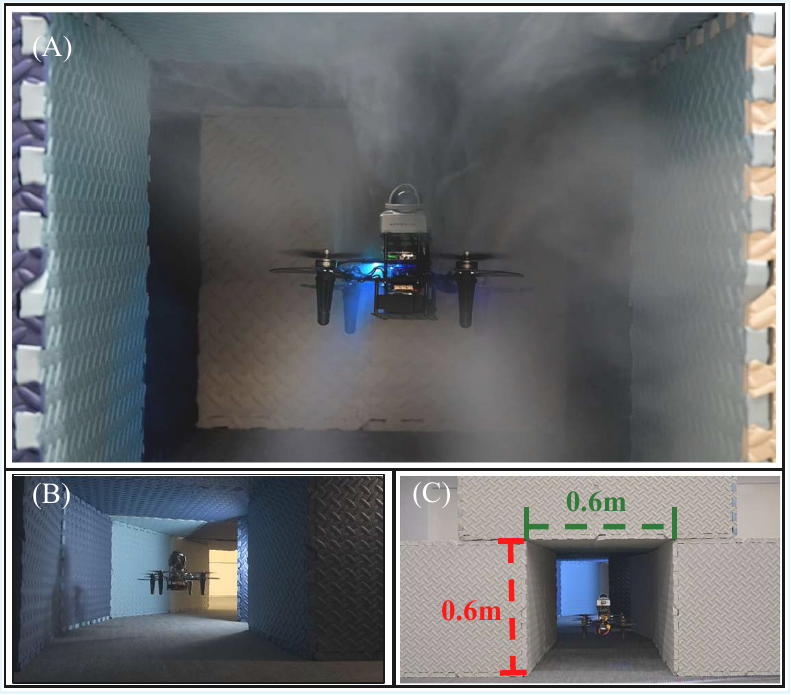}
	\caption{\textcolor{black}{Autonomous flight of the aerial robot in narrow tunnels.
			(A) The airflow disturbances are visualized using smoke.
			(B) Inside view of the tunnel.
			(C) Front view of the tunnel.}
			The video is available at
			{https://sojustfish.github.io/Disturbance-Aware-Flight-for-Aerial-Robots-in-Narrow-Space/}.
			}
	\label{disturbanceall}
\end{figure}

Existing research on autonomous flight of aerial robots in narrow environments mainly addresses these challenges from two perspectives: disturbance-aware control and motion planning. Disturbance-aware control has been studied through disturbance rejection and disturbance compensation~\cite{yu2021safety}. Disturbance compensation methods, in contrast, employ disturbance observers or incremental nonlinear dynamic inversion to explicitly estimate and compensate external disturbances~\cite{falanga2020aggressive, sun2021nmpc}. Another line of research focuses on motion planning for safe navigation. Integrated planning and tracking frameworks have been well studied for unmanned ground vehicles~\cite{zhou2021ugv}, and are further extended to aerial robots. Trajectory optimization and obstacle avoidance algorithms generate collision-free paths with sufficient geometric clearance~\cite{oleynikova2018volumetric}. In narrow tunnels, planners typically guide aerial robots along centerlines to maximize safety margins and enhance perception stability~\cite{wang2022neither}.

Despite these advances, key limitations remain. In narrow tunnels, both excessively high and excessively low flight speeds may lead to failures due to coupled aerodynamic disturbances, perception degradation, and constrained space. Existing systems often rely on offline-determined fixed speeds, which cannot adapt to varying tunnel geometries or disturbance conditions~\cite{wang2022neither}. Furthermore, disturbance estimation is typically confined to the control layer and rarely incorporated into motion planning inside the tunnel space~\cite{seo2023real}. Consequently, existing methods may result in either overly conservative or unsafe behaviors, especially in disturbance-intensive regions such as tunnel contractions.

To overcome these limitations, this paper proposes a disturbance-aware planning and control framework (DAPCF) for aerial robots operating in narrow space as shown in Fig.~\ref{disturbanceall}. The key idea is to explicitly incorporate online disturbance estimation into the planning--control loop, enabling the planner and controller to adapt flight behavior according to the environmental disturbance level and spatial constraints. The main contributions of this work are summarized as follows:

\begin{itemize}
	
	\item A disturbance-aware speed adaptation strategy that maps estimated disturbance into a risk function, enabling adaptive multi-level speed scheduling for autonomous flight in tunnels with varying confinement levels.
	
	\item A DAPCF that integrates motion planning with disturbance-compensated control using dual-loop observers and a motor-dynamics-based nonlinear model predictive controller (MDNMPC).
	
	\item Extensive real-world experiments in three representative tunnels demonstrate that a 0.39~m wide quadrotor can achieve stable autonomous flight at speeds up to 1.0~m/s inside tunnels as narrow as 0.6~m, outperforming human pilots in both success rate and flight efficiency.
	
\end{itemize}

The remainder of this paper is organized as follows. Section~\ref{Related Work} reviews related research on disturbance-aware control and autonomous flight in narrow space. Section~\ref{System Framework} presents the quadrotor dynamics and the proposed DAPCF. Real-world experimental validations are presented in Section~\ref{Real-World Experiments}. Finally, concluding remarks are given in Section~\ref{Conclusion}.

\section{RELATED WORK}
\label{Related Work}

\subsection{Autonomous Flight in Narrow and Confined Space}
Autonomous flight inside narrow space such as pipes, tunnels, and air ducts is strongly influenced by self-induced airflow recirculation and near-wall aerodynamic effects, which can introduce spatially varying disturbance forces and torques \cite{bauersfeld2025low, wang2025autonomous, thomas2025flying}. Computational fluid dynamics analysis \cite{wang2025autonomous} further suggests a characteristic speed trade-off in these environments: flying too slowly may increase the impact of recirculation and degrade stability, whereas flying too fast can challenge the limits of perception and control. Recent studies have reported autonomous traversal in confined conduits \cite{wang2025autonomous}, real-time flow sensing via event-based smoke velocimetry \cite{bauersfeld2025low}, disturbance-aware control constraints \cite{fan2024flying}, and aerodynamic force-field mapping inside air ducts \cite{thomas2025flying}. Nonlinear model predictive control (NMPC) has also been widely adopted for trajectory tracking under such challenging conditions \cite{zhou2026rotor}. Safety control methods based on flight-envelope protection and reference command generation
can reshape reference commands to keep the vehicle within admissible operating envelopes~\cite{yu2022safety}. Despite these advances, disturbance-related information is most often leveraged within the estimation/control stack, while its potential role in shaping motion planning---especially speed selection---is less explored. More specifically, existing systems for narrow space typically do not jointly account for both obstacle proximity (local geometry) and the magnitude of perceived disturbances when choosing flight speed. Some planning methods \cite{wang2022neither} rely on offline Computational Fluid Dynamics, which can be informative but may be difficult to adapt to changing conditions online. On the other hand, data-driven control strategies \cite{torrente2021data} have demonstrated promising performance, yet their deployment often benefits from substantial simulation coverage and careful domain transfer. As a result, enabling an integrated planner that adapts flight speed according to local clearance and online disturbance estimates remains an open and practically important problem for safe and efficient navigation in narrow environments.

\subsection{Disturbance Estimation and Compensation for Aerial Robots}
Accurate online estimation of aerodynamic disturbances is an important enabler for disturbance-aware flight, and a variety of approaches have been proposed. HDVIO \cite{cioffi2023hdvio, cioffi2025hdvio2} augments point-mass dynamics with learned residuals to jointly estimate states and external forces. VIMO \cite{nisar2019vimo} and VID-Fusion \cite{ding2021vid} incorporate thrust or dynamics models into the VIO factor graph. Indoor ego airflow disturbances are estimated in \cite{wang2021estimation}, although the treatment of motor-level asymmetry can be further refined. Overall, these methods provide increasingly informative disturbance estimates, which are primarily used to improve state estimation or stabilize tracking. Complementary to estimation, disturbance compensation at the control level has also been extensively studied. Incremental nonlinear dynamic inversion (INDI) enables near-model-free attitude control using high-rate sensor increments \cite{smeur2016adaptive}. A unified external-wrench estimation framework is presented in \cite{mckinnon2020estimating}. While these control-level techniques effectively mitigate disturbances within the low-level loop, the interaction between disturbance awareness and high-level planning decisions (e.g., speed scheduling in proximity to walls or contractions) is still not routinely incorporated. Recent work has explored disturbance-aware resilient planning and control for quadrotors~\cite{zhou2024internal}, but it does not tailor speed adaptation strategies for wall-induced aerodynamic effects in narrow space. Bridging this gap---using online disturbance estimates together with obstacle distance to modulate the planned flight speed---is therefore a natural next step toward more adaptive and reliable autonomy in narrow and disturbance-intensive environments.

\section{SYSTEM FRAMEWORK}
\label{System Framework}

\begin{figure}
	\centering
	\includegraphics[width=0.990\linewidth]{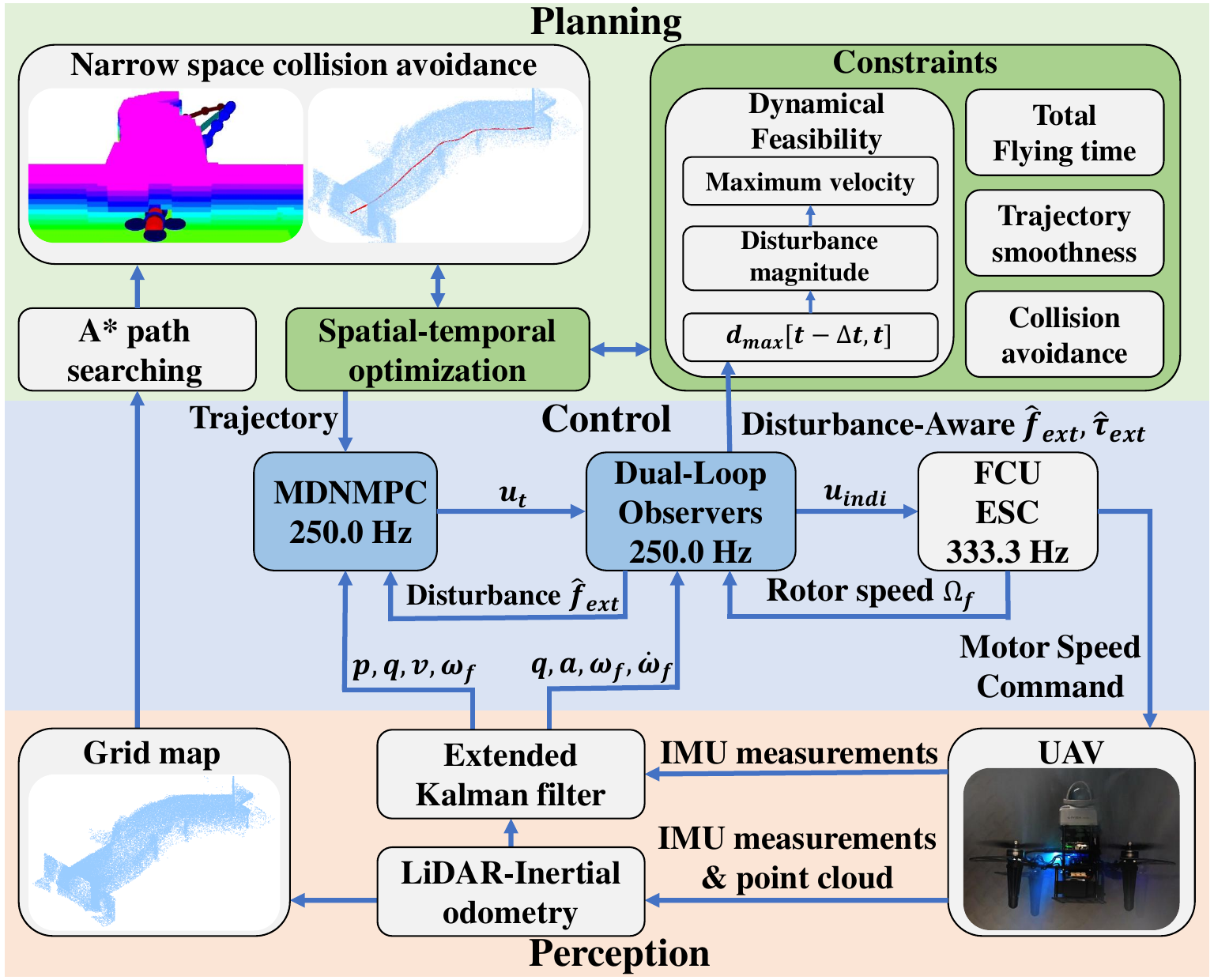}
	\caption{\textcolor{black}{The overall system framework in real-world experiments. }}
	\label{flowchart}
\end{figure}

In this section, the quadrotor dynamics model is provided for the foundation of the proposed autonomous flight system. 
Building upon this model, the section introduces the overall system framework, which integrates the disturbance-aware 
controller with the motion planner to enable safe and efficient navigation in narrow environments as shown in Fig.~\ref{flowchart}.
\subsection{Quadrotor Dynamics}
The quadrotor is modeled using rigid-body kinematics and dynamics as follows:
\begin{equation}
	m\ddot{\bm{\xi}} =
	T \bm{z}_B
	- \bm{R}(\bm{q})\bm{\Gamma}
\bm{R}^\top(\bm{q})\bm{v}
	+ m \bm{g}
	+ \bm{f}_{\text{ext}},
	\label{kinematic_model}
\end{equation}
\begin{equation}
	\dot{\bm{q}} =
	\frac{1}{2}\bm{q}\otimes
	\begin{bmatrix}
		0\\
		\bm{\omega}_B
	\end{bmatrix},
	\label{dyna_model}
\end{equation}
\begin{equation}
	\bm{J}\dot{\bm{\omega}}_B
	=
	\bm{\tau}
	-
	[\bm{\omega}_B]_\times \bm{J}\bm{\omega}_B
	+
	\bm{\tau}_{\text{ext}},
	\label{torque_model}
\end{equation}
where $\bm{\xi}$ and $\bm{v}$ denote the position and velocity of the UAV, respectively; $T$ is the total thrust generated by the rotors; $\bm{z}_B$ is the body-frame $z$-axis expressed in the world frame; $\bm{\Gamma} = diag(0.48, 0.50, 0.65) kg/s$ is the rotor-drag coefficient matrix; $\bm{R}(\bm{q})$ is the rotation matrix corresponding to the attitude quaternion $\bm{q}$; $m$ is the vehicle mass; and $\bm{g}$ denotes the gravitational acceleration. The vector $\bm{\omega}_B=[p,q,r]^\top$ represents the body angular velocity, $\bm{J}$ is the inertia matrix, and $\bm{\tau}=[\tau_x,\tau_y,\tau_z]^\top$ denotes the control torque. The terms $\bm{f}_{\text{ext}}$ and $\bm{\tau}_{\text{ext}}$ represent the external disturbance force and torque, respectively. The operator $\otimes$ denotes quaternion multiplication, and the notation $[\cdot]_\times$ denotes the skew-symmetric matrix of a vector.

The total thrust and control torques generated by the four rotors can be written as
\begin{equation}
	\begin{bmatrix}
		T \\
		\bm{\tau}
	\end{bmatrix}
	=
	\bm{M}_t \bm{t},
\end{equation}
where $\bm{t} = [T_0, T_1, T_2, T_3]^\top$ denotes the thrust vector of the four rotors. 
The thrust produced by each rotor is modeled as
\begin{equation}
	T_i = k_n \Omega_i^2,
\end{equation}
where $k_n = 1.51e^{-8} N$ is the propeller thrust coefficient and $\Omega_i$ is the angular velocity of the $i$-th rotor. The control effectiveness matrix $\bm{M}_t$ is defined as
\begin{equation}
	\bm{M}_t =
	\begin{bmatrix}
		1 & 1 & 1 & 1 \\
		-\frac{\sqrt{2}}{4}l & \frac{\sqrt{2}}{4}l & \frac{\sqrt{2}}{4}l & -\frac{\sqrt{2}}{4}l \\
		-\frac{\sqrt{2}}{4}l & \frac{\sqrt{2}}{4}l & -\frac{\sqrt{2}}{4}l & \frac{\sqrt{2}}{4}l \\
		-k_t & -k_t & k_t & k_t
	\end{bmatrix},
\end{equation}
where $l$ denotes the arm length of the quadrotor (i.e., the distance from the vehicle center to each rotor), and $k_t$ is the yaw torque coefficient.

To capture actuator dynamics and motor response delay, the rotor thrust dynamics are modeled as a first-order system:
\begin{equation}
	\dot{T}_i = \frac{1}{\sigma}(u_i - T_i),
\end{equation}
where $\sigma$ is the motor time constant obtained from experimental identification, and $u_i$ represents the commanded thrust input.

\subsection{Disturbance-Aware Controller and Observers Design}
\label{sec:nmpc_indi}
Autonomous flight inside narrow tunnels introduces strong aerodynamic proximity effects caused by surrounding walls and contraction sections. These effects generate significant unknown forces and torques that cannot be accurately modeled offline. Therefore, a disturbance-aware control architecture is adopted, consisting of an MDNMPC-based controller for trajectory tracking,  an outer-loop INDI estimating external disturbance forces, and an inner-loop INDI estimating disturbance torques. The estimated disturbances are explicitly incorporated into the predictive model and control allocation.

\subsubsection{Controller Formulation}
The system state is defined as
\begin{equation}
	\bm{x} =
	[\bm{\xi},\bm{q},\bm{v},
	\bm{\omega}_B,\bm{t}]^\top,
\end{equation}
and the control input as
\begin{equation}
	\bm{u}=[u_1,u_2,u_3,u_4]^\top .
\end{equation}
The discrete-time dynamics are
\begin{equation}
	\bm{x}_{k+1}=f(\bm{x}_k,\bm{u}_k,\hat{\bm{f}}_{\mathrm{ext}}),
\end{equation}
where $\hat{\bm{f}}_{\mathrm{ext}}$ denotes the estimated external disturbance force. The MDNMPC optimization over horizon $N$ is formulated as
{
\color{black}
\begin{equation}
	\begin{aligned}
		\bm{u}^*_{k:k+N-1}=&\arg\min_{\bm{u}} \;
		\bm{y}_{k+N}^\top \bm{Q}_N \bm{y}_{k+N}\\
		&+\sum_{i=k}^{k+N-1}
		(\bm{y}_i^\top \bm{Q}\bm{y}_i
		+\bm{u}_i^\top \bm{Q}_u\bm{u}_i) \\
		\text{s.t. }&
		\bm{x}_{i+1}=f(\bm{x}_i,\bm{u}_i,	{\bm{f}}_{\mathrm{ext},i}),\\
		&\underline{\bm{u}}\le\bm{u}_i\le\overline{\bm{u}},\\
		&\underline{{r}}\le{r}\le\overline{{r}},
	\end{aligned}
\end{equation}
where $\underline{\bm{u}}$, $\overline{\bm{u}}$, $\underline{{r}}$, and $\overline{{r}} $ represent the lower and upper bounds of control inputs, and yaw rate.

{
\color{black}
We adopt a slow-varying disturbance assumption: the estimated disturbance is held constant over the entire prediction horizon, which is valid for wall-induced aerodynamic effects in tunnels.
}

The cost vector is
\begin{equation}
	\bm{y}_i=
	\begin{bmatrix}
		\bm{\xi}-\bm{\xi}_{ref}\\
		\tilde{\bm{q}}_e\\
		\bm{v}-\bm{v}_{ref}\\
		\bm{\omega}_B-\bm{\omega}_{ref}\\
		\bm{t}-\bm{t}_{ref}
	\end{bmatrix},
\end{equation}
}where the quaternion error is $\tilde{\bm{q}_e} = \bm{q}_{ref}\otimes \bm{q}^{-1}$.
The weighting matrices are
\begin{equation}
	\bm{Q}=\mathrm{diag}
	(\bm{Q}_\xi,\bm{Q}_q,\bm{Q}_v,\bm{Q}_\omega,\bm{Q}_t),
	\quad
	\bm{Q}_N=\bm{Q}.
\end{equation}

\subsubsection{Dual-Loop Observers Design}

To account for strong aerodynamic disturbances in tunnel environments, an outer INDI-based observer estimates the external force using IMU measurements and rotor feedback.
IMU acceleration and rotor speeds are measured by the low-pass filtered as $\ddot{\bm{\xi}}_f$ and $\bm{\Omega}_f$, respectively. The total thrust is estimated as
\begin{equation}
	\hat{T}=k_n \bm{\Omega}_f^\top\bm{\Omega}_f.
\end{equation}
Rotor drag is estimated as
\begin{equation}
	\hat{\bm{f}}_{drag}
	=\bm{R}(\bm{q})\bm{\Gamma}
\bm{R}^\top(\bm{q})\bm{v},
\end{equation}
where $\bm{\Gamma}
=\mathrm{diag}(k_{d,x},k_{d,y},k_{d,z})$.
The external disturbance force is obtained by rearranging translational dynamics:
\begin{equation}
	\hat{\bm{f}}_{ext}
	=m\ddot{\bm{\xi}}_f
	-\hat{T}\bm{z}_B
	+\hat{\bm{f}}_{drag}-m \bm{g}.
	\label{outer}
\end{equation}
The estimated disturbance is injected into the MDNMPC prediction model, which enables disturbance-aware prediction during trajectory tracking.

Continuous aerodynamic disturbances inside tunnels introduce large torque uncertainties. 
Therefore, another inner INDI observer is adopted as the 
low-level attitude controller.
Following~\cite{tal2021accurate}, disturbance torque is estimated as
\begin{equation}
	\hat{\bm{\tau}}_{ext}
	=-\bm{\tau}_f
	+\bm{J}\dot{\bm{\omega}}_f
	+[\bm{\omega}_f]_\times \bm{J}\bm{\omega}_f,
	\label{inner}
\end{equation}
where $\bm{\omega}_f$ and $\dot{\bm{\omega}}_f$ are filtered angular velocity and acceleration.
The desired torque becomes
\begin{equation}
	\bm{\tau}_d
	=\bm{\tau}_f
	+\bm{J}\dot{\bm{\omega}}_d
	-\bm{J}\dot{\bm{\omega}}_f.
\end{equation}
where $\bm{\omega}_d$ is the desired angular velocity.
Finally, rotor thrust commands are computed via control allocation:
\begin{equation}
	\bm{u}_{indi}
	=\bm{M}_t^{-1}
	\begin{bmatrix}
		T\\
		\bm{\tau}_d
	\end{bmatrix},
\end{equation}
where $\bm{M}_t^{-1}$ denotes the Moore-Penrose pseudoinverse of the control effectiveness matrix. The combined inner-outer INDI observers allow simultaneous estimation of disturbance forces and torques, enabling robust and adaptive flight performance in disturbance-intensive tunnel environments.

\definecolor{mygray}{gray}{.9}

\subsection{Disturbance-Aware Trajectory Planning}
\label{Disturbance_aware_planning}

{
\color{black}
In narrow environments, collision risks arise from geometric constraints, disturbance-induced errors, and limited deceleration. We propose a disturbance-aware planner with front-end path generation (dynamic A*) and back-end optimization for smoothness and feasibility using real-time disturbance measurements to adjust velocity limits. Adaptive trajectory generation occurs upon significant disturbance changes or collision detection. When disturbances vary, the trajectory updates velocity limits while preserving path topology; otherwise, replanning is triggered only by collisions. Obstacles remain quasi-static during each replanning interval.

The back-end refines the trajectory by constructing a continuous polynomial $\mathbf{r}(t) \in \mathbb{R}^n$. This $n$-dimensional trajectory is a piecewise polynomial consisting of $M$ segments. Each segment is a quintic polynomial of degree $D = 2k - 1$, with $k = 3$. The trajectory is encoded in a coefficient matrix $\mathbf{H} \in \mathbb{R}^{2Mk \times n}$, which is derived from the via-points $\mathbf{W} \in \mathbb{R}^{n \times (M-1)}$ and the duration vector $\boldsymbol{\Delta} = [\Delta_1, \dots, \Delta_M]^\top \in \mathbb{R}_{>0}^{M}$. 
Let $t_g$ denote the cumulative time up to the end of the $g$-th segment, i.e., $t_g = \sum_{j=1}^{g} \Delta_j$ with $t_0 = 0$. The segment trajectory is defined as:
\begin{equation}
	\mathbf{r}(t) = \mathbf{r}_g(t - t_{g-1}), \quad t \in [t_{g-1}, t_g].
\end{equation}

The full trajectory is obtained by concatenating the segments:
\begin{equation}
	\mathbf{r}_g(t) = \mathbf{H}_g^\top \boldsymbol{\phi}(t), \quad t \in [0, \Delta_g],
\end{equation}
where $\boldsymbol{\phi}(t) = [1, t, \dots, t^D]^\top$ is the polynomial basis.

To generate a high-quality trajectory, we formulate a joint optimization problem 
that balances time cost, smoothness, dynamic feasibility, and collision avoidance:
\begin{equation}
	\min_{\mathbf{W}, \boldsymbol{\Delta}} J = \lambda_t J_t + \lambda_s J_s + \lambda_d J_d + \lambda_c J_c,
\end{equation}
where the time cost $J_t$ and smoothness penalty $J_s$ are defined as:
\begin{equation}
	J_t = \sum_{g=1}^{M} \Delta_g,\quad J_s = \int_{t_0}^{t_M} \|\mathbf{r}^{(k)}(t)\|_2^2 ,dt,
\end{equation}

Our planner incorporates online disturbance estimates for adaptive speed regulation.
The scalar disturbance intensity is defined as a weighted norm of the estimated force and torque disturbances:
\begin{equation}
	d(t) = w_f \left\| \hat{\bm{f}}_{\mathrm{ext}}(t) \right\| + w_{\tau} \left\| \hat{\bm{\tau}}_{\mathrm{ext}}(t) \right\|
	\label{eq:disturbance_intensity}
\end{equation}
where $w_f$ and $w_{\tau}$ are weighting coefficients.
Higher disturbance intensity implies more severe aerodynamic conditions, requiring more conservative velocity constraints.

To suppress transient fluctuations, we take the peak value of $d(t)$ over a sliding window as the effective disturbance $\hat{d}_p(t)$ for speed adjustment.
The maximum allowable velocity is then dynamically bounded as
\begin{equation}
	v_{\mathrm{max}}(t) = F\left(v_{\mathrm{nom}} - \alpha_v \hat{d}_p(t),\ v_{\mathrm{min}},\ v_{\mathrm{nom}}\right),
\end{equation}
where $F(b_0, b_1, b_2)$ bounds $b_0$ to the interval $[b_1, b_2]$, i.e., $F(b_0, b_1, b_2) = \max(b_1, \min(b_0, b_2))$. 
Here \(\hat{d}_{\mathrm{p}}(t)\) is the maximum \(\hat{d}_{\mathrm{p}}(\tau)\) of over the past \( \Delta T \) seconds. 
The nominal speed \( v_{\mathrm{nom}} \) is reduced proportionally to the peak disturbance, with sensitivity \( \alpha_v \), 
ensuring a safety margin under persistent or large disturbances.

Dynamic feasibility is enforced by penalizing kinematic violations at discrete time steps $\{ t_i \}_{i=0}^{k}$:
\begin{equation}
	J_d = J_{d,v} + J_{d,a} + J_{d,j},
\end{equation}
where the velocity, acceleration, and jerk penalties are computed as
\begin{equation}
	J_{d,v} = \sum_{i=0}^{k} \max\left\{\|\dot{\mathbf{r}}(t_i)\|_2^2 - v_{\max}(t_i)^2, 0\right\},
\end{equation}
\begin{equation}
	J_{d,a} = \sum_{i=0}^{k} \max\left\{\|\ddot{\mathbf{r}}(t_i)\|_2^2 - a_{\max}^2, 0\right\},
\end{equation}
\begin{equation}
	J_{d,j} = \sum_{i=0}^{k} \max\left\{\|\dddot{\mathbf{r}}(t_i)\|_2^2 - j_{\max}^2, 0\right\}.
\end{equation}

Obstacle surfaces are approximated as hyperplanes defined by $(\mathbf{x}^o - \mathbf{s}^o)^\top \mathbf{n}^o = 0$, where $\mathbf{n}^o$ is the outward normal pointing toward free space. The collision cost penalizes incursions into the safety margin $C_o$:
\begin{equation}
	J_c = \sum_{i=0}^{k} \max\left\{C_o - d_o(\mathbf{r}(t_i)), 0\right\}^2,
\end{equation}
with $d_o(\mathbf{r}(t_i))$ the signed distance from the trajectory point to the nearest obstacle. By dynamically adapting the trajectory to disturbances and collision risks, the planner enables safe and agile navigation in narrow space while preserving the topological structure of the initial path.

\section{REAL-WORLD EXPERIMENTS}
\label{Real-World Experiments}

The effectiveness of the proposed disturbance-aware control and planning framework is validated through a series of real-world flight experiments. In this section, we aim to answer three key questions:

\begin{enumerate}
	\item Can the dual-loop observers achieve accurate disturbance estimation, and does the MDNMPC maintain tracking stability under uncertainties? (Refer to Section~\ref{Controller Ablation Study}.)
	\item Can the planner adapt velocity according to varying confinement levels in narrow space? (Refer to~\ref{Disturbance-Aware Planning Performance}.)
	\item How does the proposed disturbance-aware flight system compare with different commercial UAVs? (Refer to~\ref{Tunnel Comparison Results}.)
\end{enumerate}

The proposed system is developed within the ROS framework, with inter-module communication relying on the MAVROS protocol~\cite{Mavros2024git}.
Perception, planning, and control algorithms run on an onboard computer (NVIDIA Jetson Orin NX). Desired motor speeds are computed and sent 
to a Kakute H7 flight controller running the ArduPilot firmware, which handles low-level motor speed control.
The physical parameters of the quadrotor platform are listed in Table~\ref{table_1}. 
In the MDNMPC formulation, the prediction horizon \(N\) is set to 20, and the sampling 
interval \(dt\) is 50\,\(ms\). A complete list of MDNMPC parameters is provided in Table~\ref{table_2}.
}

\begin{table}[!t]
	\renewcommand{\arraystretch}{2}
	\caption{Parameters of the Quadrotor Dynamics}
	\label{table_1}
	\centering
	\begin{tabular}{llll}
		\toprule
		Parameter & Value & Parameter & Value \\
		\midrule
	\rowcolor{mygray}	$m$ [$kg$] & 1.05 & 	 $\bm{J}$ [$kgm^{2}$] & diag(4.77, 5.26, 5.44)$e^{-3}$\\
	 $k_{t}$ & 0.015 & $\sigma$ [$s$] & 0.035 \\
		\rowcolor{mygray}	$l$ [$m$] & $0.25$ & $k_{n}$ & 1.51$e^{-8}$ \\
		\bottomrule
	\end{tabular}
\end{table}

\begin{table}[!t]
 	\renewcommand{\arraystretch}{1.5} 
 	\caption{Parameters Selection for Control System}
 	\label{table_2}
 	\centering
 	\begin{tabular}{llll}
 		\toprule
 		Parameter& Value & Parameter & Value \\
 		\midrule
 		$ \bm{Q}_{p} $ & $\mathrm{diag}(300, 300, 800)$ &	$ \bm{Q}_{v} $ & $\mathrm{diag}(1, 1, 1)$ \\
 		\rowcolor{mygray} $ \bm{Q}_{q} $ & $\mathrm{diag}(20, 20, 20, 20)$ & 	$ \bm{Q}_t $ & $\mathrm{diag}(0.5, 0.5, 0.5)$ \\
 		$ \bm{Q}_{\omega} $ & $\mathrm{diag}(1, 1, 1)$ &  $ \bm{Q}_u $ & $\mathrm{diag}(1, 1, 1, 1)$ \\
 		
 		\bottomrule
 	\end{tabular}
\end{table}

\subsection{Controller Ablation Study}
\label{Controller Ablation Study}

\begin{figure}
	\centering
	\includegraphics[width=0.990\linewidth]{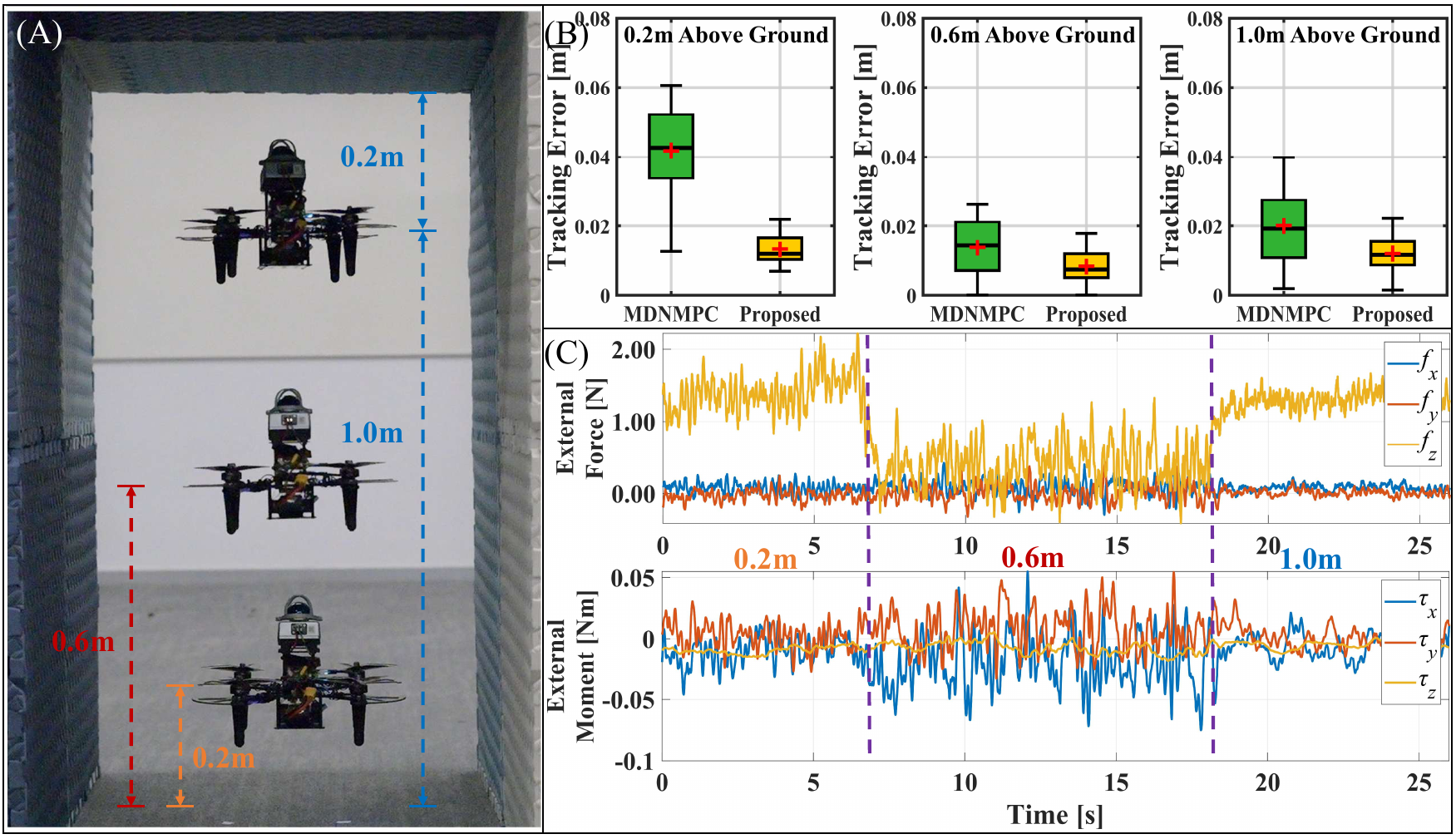}
	\caption{\textcolor{black}{Experimental results of the proposed method and MDNMPC. 
	(A) The aerial robot autonomously hovering at heights of 0.2 m, 0.6 m, and 1.0 m in a narrow space.
	(B) Comparison of the MDNMPC and the proposed method at different heights. 
	(C) External force disturbances and external moment disturbances estimated by INDI.}}
	\label{hovering}
\end{figure}

\begin{figure*}
	\centering
	\includegraphics[width=0.990\linewidth]{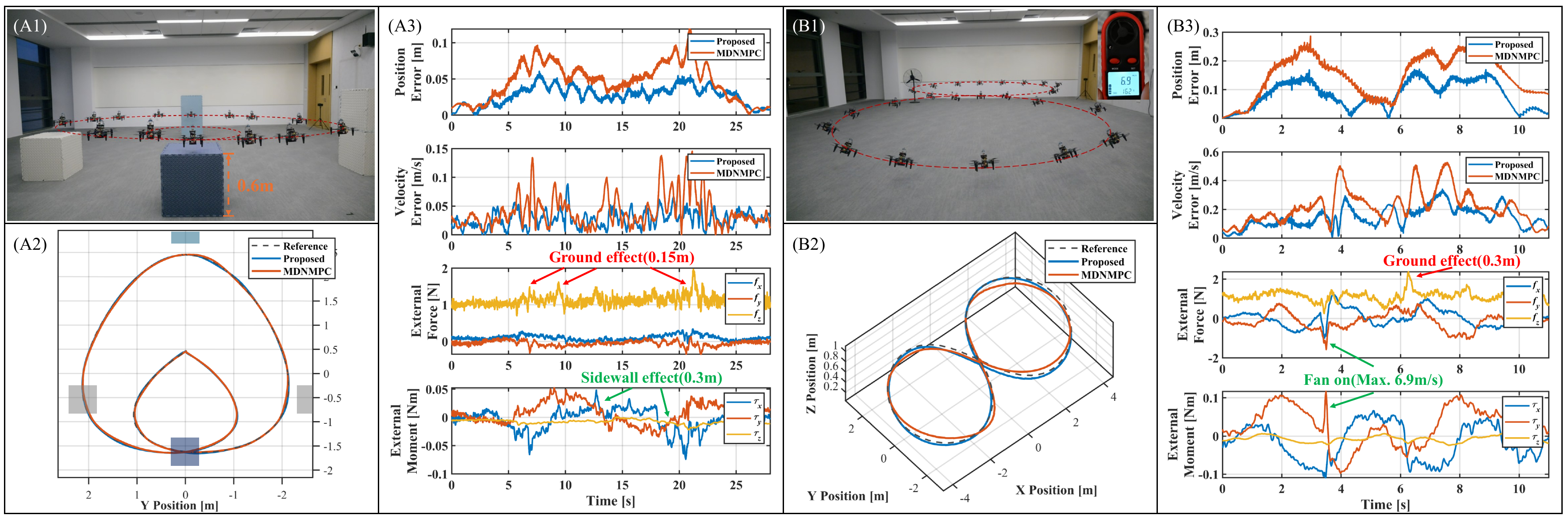}
	\caption{\textcolor{black}{ Disturbance-aware trajectory tracking. 
	(A1) Overlaid snapshots of the aerial robot autonomously following a circular trajectory. 
	(A2) Tracking performance of the circular trajectory under two controllers.
	(A3) The tracking error and estimated disturbance under the proposed control system.
	(B1) Overlaid snapshots of the aerial robot autonomously following a lemniscate trajectory. 
	(B2) Tracking performance of the lemniscate trajectory under two controllers.
	(B3) The tracking error and estimated disturbance under the proposed control system.
	}}
	\label{tracking}
\end{figure*}
{
\color{black}
To evaluate disturbance rejection performance, we conducted three hovering tests at 0.2 m, 0.6 m, and 1.0 m above ground, all 0.3 m from a sidewall (Fig. \ref{hovering}(A)). These heights represent ground-effect, mid-altitude, and ceiling-effect conditions with distinct aerodynamic disturbance profiles.

Fig. \ref{hovering}(C) shows the estimated disturbance forces and torques. The mean vertical disturbance force is ~1.4 N under ground effect, ~1.3 N under ceiling effect, and only ~0.5 N at 0.6 m where boundary interactions are weak, confirming that surface proximity significantly amplifies aerodynamic disturbances.

Tracking performance is compared in Fig. \ref{hovering}(B). The baseline MDNMPC yields mean position errors of 0.042 m, 0.014 m, and 0.020 m at the three heights, while the proposed method reduces them to 0.013 m, 0.008 m, and 0.012 m. It consistently achieves lower tracking error and smaller variance, with the most pronounced improvement under strong ground and ceiling effects.
}

Additionally, we carried out two sets of experiments to evaluate the disturbance-aware trajectory tracking performance of the aerial robot under ground/sidewall
effects, agile maneuvers, and external wind disturbances. These include
circular and lemniscate trajectories, with comparisons made between INDI-enabled and INDI-disabled modes.
\begin{enumerate}
	\item  Circular Trajectory Test: Two $4\,\mathrm{m} \times 4\,\mathrm{m}$ circular trajectories
	are designed, with a fixed flight height of $0.85\,\mathrm{m}$. The aerial robot takes off at
	$0.10\,\mathrm{m}$; $0.6\,\mathrm{m}$-high boxes are placed on (inducing ground effect) and
	beside (inducing sidewall effect) the trajectories, maintaining a $0.15\,\mathrm{m}$ height
	difference between the box top and trajectory (see Fig.~\ref{tracking}(A1)). Flight parameters: velocity $1.0\,\mathrm{m/s}$,
	acceleration $4.0\,\mathrm{m/s^2}$. Four boxes are placed to enhance disturbances.
	\item Lemniscate Trajectory Test: An $8\,\mathrm{m} \times 4\,\mathrm{m}$
	variable-altitude lemniscate trajectory is designed, starting at 0.6 m, with a maximum height
	of $0.9\,\mathrm{m}$ and minimum height of $0.3\,\mathrm{m}$ (to extend low-altitude flight time
	and enhance ground effect), illustrated in Fig.~\ref{tracking}(B1). Flight parameters: velocity $5.0\,\mathrm{m/s}$, acceleration
	$8.0\,\mathrm{m/s^2}$ (to simulate agile maneuvers); a fan is placed along the trajectory to generate wind disturbance,
	which can achieve a wind velocity of up to $6.9\,\mathrm{m/s}$.
\end{enumerate}

Fig.~\ref{tracking} presents the trajectory tracking results and disturbance estimation data.
Fig.~\ref{tracking}(A2) and (B2) show tracking performance of the two trajectories under two controllers is compared.
For the circular trajectory in Fig.~\ref{tracking}(A3), the INDI-enabled method achieves the RMSE
of 0.030 m. Without INDI, the RMSE exceeds 0.054 m. The tracking error is reduced by 43\%.
With INDI, the maximum tracking error and velocity error are 0.060 m and 0.088 m/s, respectively.
These values are much lower than 0.117 m and 0.142 m/s obtained without INDI.
This demonstrates that INDI effectively suppresses disturbances caused by ground and sidewall effects.
For the lemniscate trajectory in Fig.~\ref{tracking}(B3), the tracking error rises to about 0.163 m without INDI.
Severe tracking degradation occurs under wind disturbances. In contrast, the INDI-enabled mode
maintains stable tracking with the RMSE of 0.095 m. This performance is achieved even under
5.0 m/s agile maneuvers, ground effect, and wind disturbances. The tracking error
is reduced by 41\%. With INDI, the maximum tracking error and velocity error are 0.171 m and 0.340 m/s, respectively.
These values are much lower than 0.286 m and 0.522 m/s obtained without INDI. During sudden disturbances
such as ground effect, sidewall effect, and strong wind,
the INDI observer estimates and compensates for complex disturbances in real time.
It ensures more robust and anti-disturbance trajectory tracking performance.

{
\color{black}

\begin{figure*}
	\centering
	\includegraphics[width=0.990\linewidth]{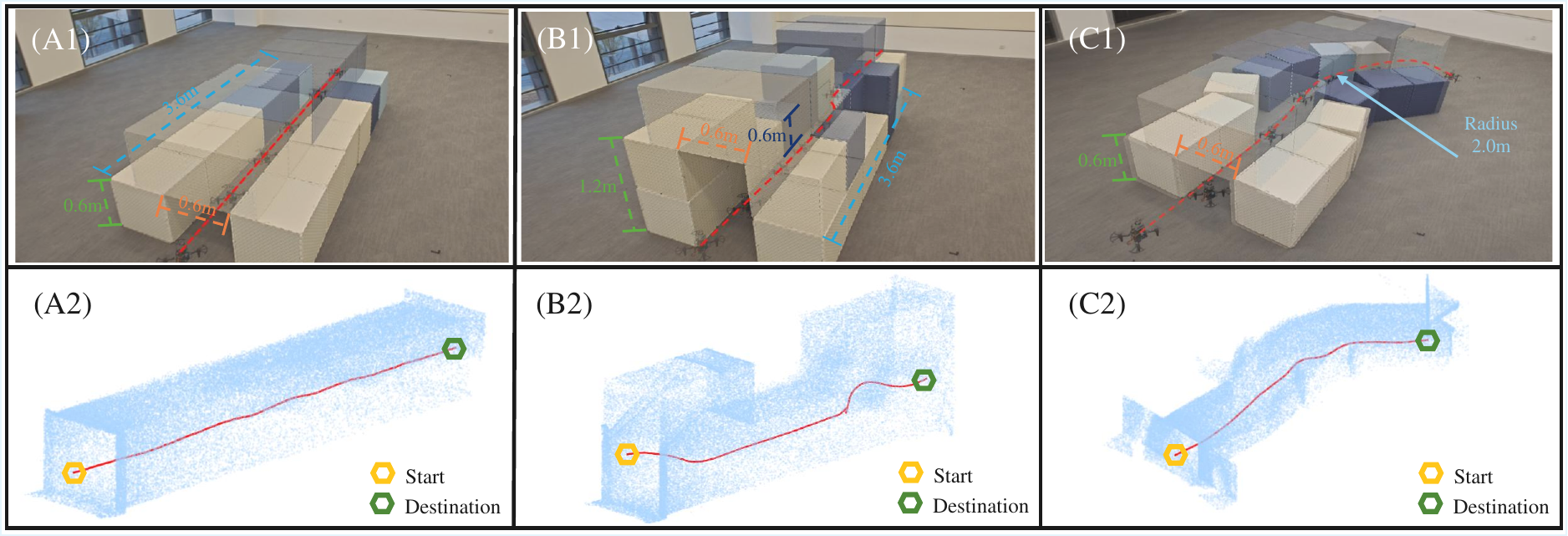}
	\caption{\textcolor{black}{ Three tunnels for validation of disturbance--aware flight.
	(A1) Snapshots of the proposed system flying through a \textbf{straight} tunnel (3.6 m length, 0.6 m width and 0.6 m height). 
	(A2) Visualization of the flight process in \textbf{straight} tunnel.
	(B1) Snapshots of the proposed system flying through a \textbf{sloped} tunnel (3.6 m length, 0.6 m width and 0.6-1.2 m height). 
	(B2) Visualization of the flight process in \textbf{sloped} tunnel.
	(C1) Snapshots of the proposed system flying through a \textbf{curved} tunnel (2.0 m radius, 0.6 m width and 0.6 m height). 
	(C2) Visualization of the flight process in \textbf{curved} tunnel.
	}}
	\label{cloud}
\end{figure*}

\begin{table}[!t]
 	\renewcommand{\arraystretch}{1.5} 
 	\caption{Experimental Results of the Proposed System in Various Tunnels}
 	\label{table_3}
 	\centering
 	\begin{tabular}{lccc}
 		\toprule
		Seq. & Straight tunnel & Sloped tunnel & Curved tunnel \\
 		\midrule
 		Avg. disturbance & 1.8007 & 1.4965 & 1.9109 \\
 		\rowcolor{mygray}Avg. Vel. [$m/s$] & 0.4744 & 0.6441 & 0.4960 \\
		Max. Vel. [$m/s$] & 0.9299 & 1.0505 & 0.7210 \\
		\rowcolor{mygray}Traj. duration [$s$] & 8.34 & 6.63 & 7.28 \\
 		\bottomrule
 	\end{tabular}
\end{table}

\begin{figure}
	\centering
	\includegraphics[width=0.90\linewidth]{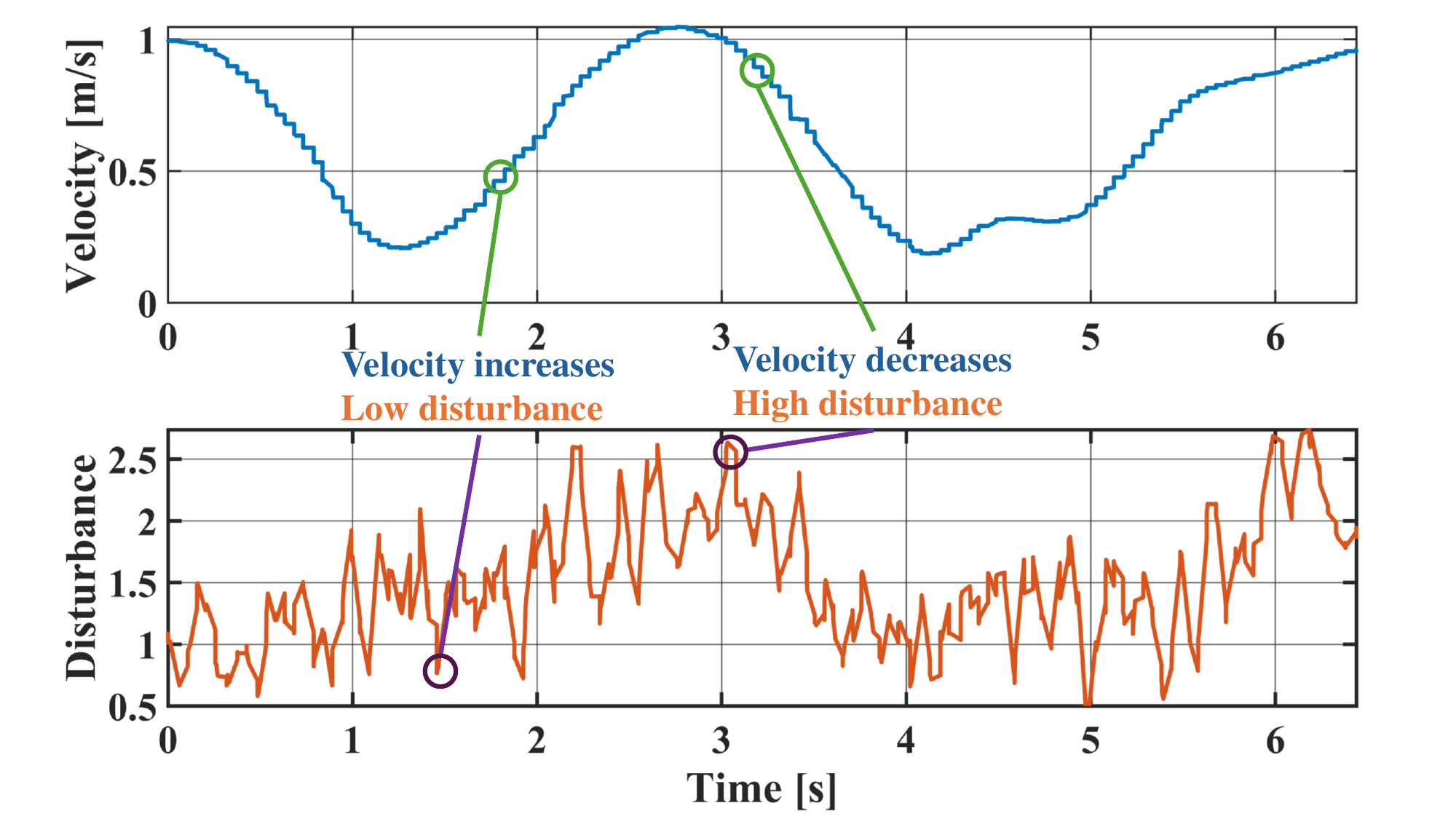}
	\caption{\textcolor{black}{ Disturbance and velocity profiles during autonomous flight in the sloped tunnel. }}
	\label{autoplanning}
\end{figure}

\subsection{Disturbance-Aware Planning Performance}
\label{Disturbance-Aware Planning Performance}

We validate the proposed planner in three distinct realistic narrow tunnels: a straight tunnel, 
a sloped tunnel, and a curved tunnel, as illustrated in Fig.~\ref{cloud}. The straight tunnel 
is 3.6 m in length, 0.6 m in width, and 0.6 m in height. Fig.~\ref{cloud}(A1) presents a sequence of snapshots 
during the flight, while Fig.~\ref{cloud}(A2) shows the point cloud of the tunnel interior acquired by the onboard 
LiDAR. The sloped tunnel is 3.6 m long and 0.6 m wide, with a height ranging from 0.6 m to 1.2 m. 
Its flight snapshot sequence and point cloud are provided in Fig.~\ref{cloud}(B1) and (B2), respectively. 
The curved tunnel has a radius of 2.0 m, a width of 0.6 m, a height of 0.6 m, and a curvature radius 
of about 2.0 m. The corresponding snapshot sequence and point cloud are displayed in Fig.~\ref{cloud}(C1) and (C2).

The proposed system completed multiple flights in all three tunnels without any failures. This demonstrates the stability and robustness of both the motion planning algorithm and the autonomous flight system in narrow tunnel environments. Table~\ref{table_3} summarizes the experimental results.

In the straight and curved tunnels, the average disturbance magnitudes were 1.8007 and 1.9109, respectively. Both values are significantly higher than the 1.4965 observed in the sloped tunnel. In contrast, the average flight velocities in the straight and curved tunnels were 0.4744 m/s and 0.4960 m/s. These are considerably lower than the 0.6441 m/s achieved in the sloped tunnel. The maximum velocities recorded in the straight, sloped, and curved tunnels were 0.9299 m/s, 1.0505 m/s, and 0.7210 m/s, respectively. The lowest maximum velocity in the curved tunnel results from the highest disturbance level combined with the need for conservative planning to accommodate continuous direction changes and tighter obstacle avoidance constraints.


In the sloped tunnel, the passage height changes continuously. This leads to noticeable fluctuations in the disturbance perceived by the aerial robot. The proposed disturbance-based velocity adjustment strategy proves effective in this environment. The aerial robot automatically decreases its velocity when disturbance increases and accelerates again when disturbance subsides. Fig.~\ref{autoplanning} shows the relationship between the actual flight velocity and the defined disturbance, demonstrating that the robot adjusts its velocity in response to varying disturbance levels to ensure safe flight.

\subsection{Tunnel Comparison Results}
\label{Tunnel Comparison Results}
We perform ablation studies on the proposed planning method and compare it with manual flights by an experienced pilot using a commercial quadrotor (DJI NEO2) and a micro system (football drone). Experiments are conducted in straight, sloped, and curved tunnel environments. The three systems—our proposed system, DJI NEO2, and the football drone—are shown in Fig.~\ref{threesystems}, with maximum widths in the y-direction of 0.39 m, 0.32 m, and 0.20 m, respectively.

\begin{figure}
	\centering
	\includegraphics[width=0.90\linewidth]{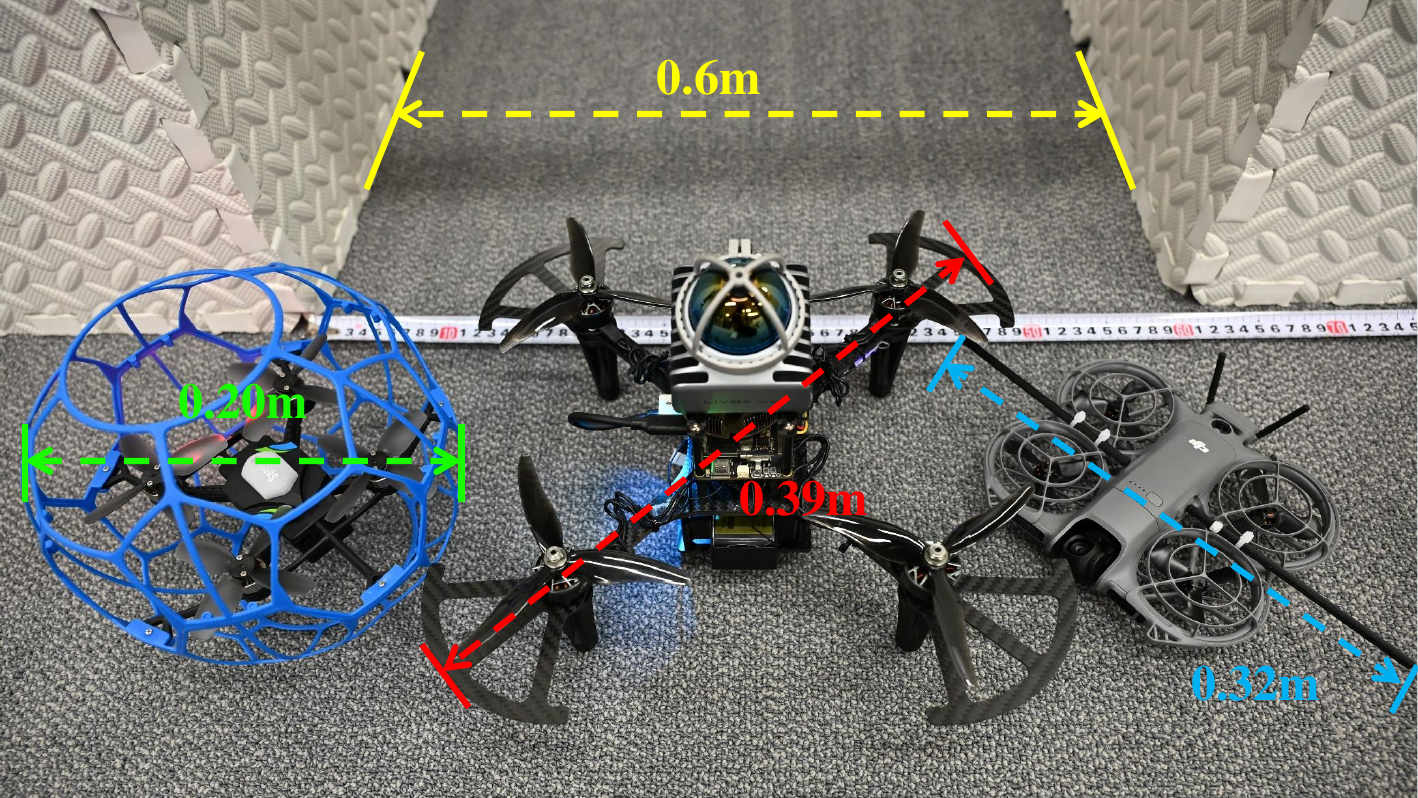}
	\caption{\textcolor{black}{The three systems evaluated in tunnel experiments: (red) proposed system (width 0.39 m), (blue) DJI NEO2 (width 0.32 m), (green) football drone (width 0.20 m). }}
	\label{threesystems}
\end{figure}

\begin{table}[!t]
 	\renewcommand{\arraystretch}{1.5} 
 	\caption{Comparison of Results Across Different Systems and Tunnels}
 	\label{table_4}
 	\centering
 	\begin{tabular}{lccc}
 		\toprule
		System & Straight tunnel & Sloped tunnel & Curved tunnel \\
 		\midrule
 		Proposed & $\circ$ & $\circ$ & $\circ$ \\
 		\rowcolor{mygray}DJI NEO2 & $\times$ & $\times$ & $\times$ \\
 		Football drone & $\times$ & $\times$ & $\times$ \\
 		\bottomrule
		\multicolumn{4}{l}{$\circ$: success, $\times$: fail.}
 	\end{tabular}
\end{table}

\begin{figure}
	\centering
	\includegraphics[width=0.990\linewidth]{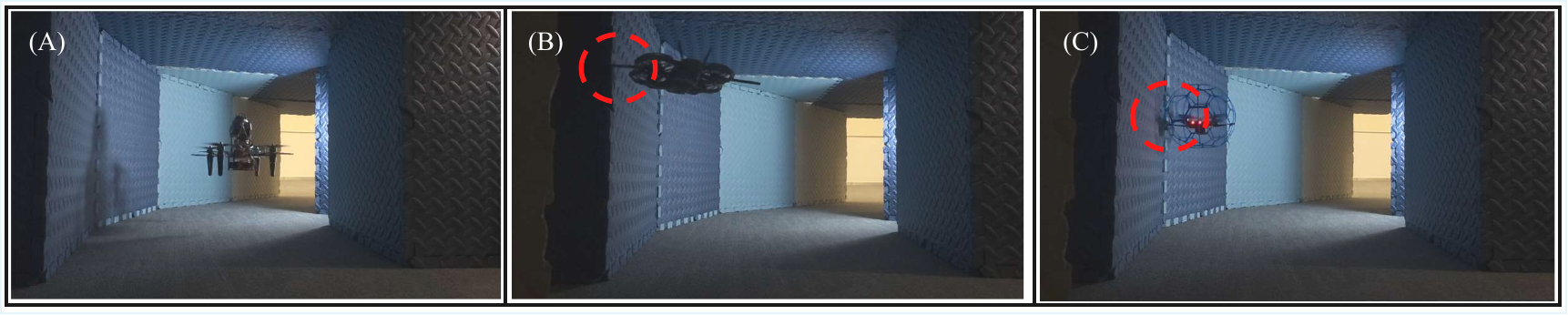}
	\caption{\textcolor{black}{Experimental comparison of the three systems in tunnel environments. 
	(A) Successful autonomous flight of our aerial robot.
	(B) DJI NEO2 collides with the tunnel wall (red circle).
	(C) The football drone collides with the tunnel wall (red circle).}}
	\label{collision}
\end{figure}

As shown in the traversal results in Table~\ref{table_4}, the proposed system successfully navigates all tunnel scenarios. In contrast, manual flights fail to complete any of the tunnel tasks. Human operators struggle to maintain real-time situational awareness in confined and visually degraded environments. Keeping the aerial robot within the safe region of the tunnel is extremely difficult. Even minor deviations tend to attract the drone toward the wall due to ground effect and airflow disturbances. Once the aerial robot drifts outside the safe region, the narrow space leaves little margin for recovery. Collisions often occur before the pilot can react. These results highlight the superior performance, stability, and reliability of the proposed system compared with experienced human pilots under challenging tunnel conditions. Fig.~\ref{collision} shows representative snapshots of successful and failed traversals for the three systems in tunnel environments.

\section{CONCLUSION}
\label{Conclusion}

This paper presented the DAPCF for autonomous quadrotor flight in narrow tunnels, featuring the dual-loop observers for real-time 6-degree-of-freedom disturbance estimation, a planner that adapts flight speed based on varying confinement levels, and a nonlinear controller incorporating motor dynamics and external disturbances. Real-world experiments validated the framework: the dual-loop observers captured distinct disturbance patterns under ground and ceiling effects, reducing tracking error by over 40\% compared to MDNMPC without compensation. In straight, sloped, and curved tunnels, the disturbance-aware system completed multiple flights without failure, with the speed regulation strategy responding to local disturbance variations by reducing speed when disturbances increased and accelerating when conditions improved. Comparative experiments showed that human pilots failed to traverse any tunnels due to limited situational awareness, confirming that integrating online disturbance estimation into both planning and control enables safe and efficient flight in narrow space.
}

\end{document}